\documentclass[aps,pre,reprint,groupedaddress,nofootinbib,floatfix]{revtex4-2}

\usepackage{amsmath,amssymb,amsfonts}
\usepackage{mathtools}
\usepackage{graphicx}
\usepackage{xcolor}    
\colorlet{RED}{red}    
\colorlet{BLUE}{blue}
\usepackage[ruled,vlined]{algorithm2e}  
\usepackage[hidelinks]{hyperref}  

\SetAlFnt{\small}
\SetAlCapFnt{\small}
\SetAlCapNameFnt{\small}

\newcommand{\diff}{\mathrm{d}}
\newcommand{\rfield}{\mathcal{R}}      
\newcommand{\Fop}{\mathcal{F}}         
\newcommand{\Lop}{\mathcal{L}}         
\newcommand{\adj}{\lambda}             

\usepackage{placeins}

\begin{document}

\title{
Transformer Field Theory\\[0.25em]
\normalsize A Response-Theoretic Approach to Mechanistic Interpretability
}

\author{David N. Olivieri}
\email{olivieri@uvigo.gal}
\affiliation{Departamento de Informática, Universidade de Vigo, Spain}

\author{Antonio F. P\'erez Rodr\'iguez}
\email{antfperez@gmail.com}
\affiliation{Independent Researcher, Spain}


\begin{abstract}
Mechanistic interpretability often studies Transformer behavior by intervening on internal activations through activation patching, causal tracing, path patching, and steering directions. This paper develops \emph{Transformer Field Theory}: a response-theoretic framework in which the residual stream of a fixed forward pass is treated as a Transformer field over layer depth and token position. In this formulation, patching becomes a localized source insertion into the Transformer field, first-order sensitivity fields predict patch effects, Green functions describe downstream propagation, and patch selection is posed as an adjoint inverse problem. Empirically, we test the theory’s forward response objects in GPT-2-style autoregressive Transformers. Localized Transformer-field interventions exhibit a bounded local linear regime; first-order sensitivities predict patch effects across layer-token sites; localized sources generate structured anisotropic Transformer-field propagation; high-sensitivity sites and sliced Green operators provide reduced response descriptions; and prompt-induced Transformer-field displacements partially transfer answer behavior. These results establish sensitivities, Transformer-field responses, and sliced Green operators as practical objects for organizing patching experiments, while providing the forward mathematical basis for patch-site inference and cross-scale response transfer.
\end{abstract}

\maketitle

\section{Introduction}
\label{sec:intro}

Transformers and large language models (LLMs) exhibit rich internal structure across layers, tokens, attention heads, and nonlinear representation subspaces.\cite{vaswani2017attention} A central goal of mechanistic interpretability is to identify the internal mechanisms responsible for model behavior by intervening on activations and measuring changes in outputs.\cite{olah2020circuits,elhage2021framework} Activation patching, causal tracing, path patching, and steering directions have shown that behaviorally meaningful sites and directions can be localized within these representations.\cite{wang2023ioi,goldowskyDill2023pathpatching,zhang2023patching} This paper asks a complementary question: can such interventions be organized and predicted by response quantities derived from a field-theoretic description of the residual stream as a depth-token field?

We study this question by treating a Transformer-based LLM as a residual dynamical system evolving along network depth. The central object is the residual stream itself. Fix a Transformer model \(M\), a prompt \(q\), and a forward pass. Write
\begin{equation}
R^M(q;\ell,x,i), \qquad i=1, \dots, d_{\rm model},
\label{eq:intro_residual_component}
\end{equation}
for the \(i\)-th residual-stream component at layer \(\ell\) and token position \(x\). Collecting the components gives the full residual-stream vector
\begin{equation}
R^M(q;\ell,x)
=
\bigl(
R^M(q;\ell,x,1),\dots,
R^M(q;\ell,x,d_{\rm model})
\bigr)
\label{eq:intro_residual_vector}
\end{equation}
with \(R^M(q;\ell,x)\in\mathbb{R}^{d_{\rm model}}\) at the depth-token site \((\ell,x)\). For a fixed model \(M\), prompt \(q\), and forward pass under consideration, we suppress the labels \(M\) and \(q\) and define the \textbf{discrete Transformer field}:
\begin{equation}
\rfield_\ell(x) :=  R^M(q;\ell,x)
\in \mathbb{R}^{d_{\rm model}} .
\label{eq:intro_discrete_transformer_field}
\end{equation}
Thus \(\rfield_\ell(x)\) denotes the Transformer-field vector at layer \(\ell\) and
token position \(x\) for the fixed forward pass.

Its corresponding continuum idealization is the \textbf{continuum Transformer field}:
\begin{equation}
\rfield(t,x)\in\mathbb{R}^{d_{\rm model}},
\qquad
(t,x)\in[0,T]\times\mathcal{X}.
\label{eq:intro_continuum_transformer_field}
\end{equation}
In this notation, the layer index \(\ell\) is represented by a continuous depth coordinate \(t\in[0,T]\), and the token position \(x\) is treated as a coordinate in the token-position domain \(\mathcal{X}\). The same symbol \(x\) is used for the discrete token position and for its continuum idealization, with the intended meaning fixed by context.

Mechanistic interventions are then localized sources acting on the continuum Transformer field of Eq.~\eqref{eq:intro_continuum_transformer_field}. Their effects are described by response functions of the continuum Transformer field. This yields a theoretical structure for patching: patch effects, Transformer-field propagation, and patch-site inference can all be formulated within the same linear-response formalism.

In a local linear regime, an intervention $J(t,x)$ induces a first-order change in an observable $y$. For example, $y$ may be a logit difference or a behavioral score. The response-theoretic form is
\begin{align}
\delta y
&\approx
\int\!\diff t \int\!\diff x\,
a(t,x)\,J(t,x),
\label{eq:intro_response}
\\[2pt]
a(t,x)
&:=
\frac{\delta y}{\delta \rfield(t,x)} .
\nonumber
\end{align}
Thus the sensitivity field $a(t,x)$ gives a first-order predictor of patch effects. Downstream propagation is described, at the formal level, by the two-point response, or Green function,
\begin{equation}
G(t,x;t',x')
:=
\frac{\delta \rfield(t,x)}
{\delta \rfield(t',x')},
\label{eq:intro_Gdef}
\end{equation}
which, for fixed source and target sites, is a $d_{\mathrm{model}}\times d_{\mathrm{model}}$ matrix. It measures how a perturbation injected at $(t',x')$ propagates through depth and across token position.

The role of \(G\) is to provide the forward operator for patch localization as
a constrained inverse problem. Given a desired Transformer-field displacement or
behavioral shift, one can seek a sparse, structured source \(J\) whose
propagated response satisfies an integral equation of the form
\begin{equation}
\begin{aligned}
\Delta\rfield_{\mathrm{target}}(t,x)
&\approx
\int\!\diff t' \int\!\diff x'\,
G(t,x;t',x')\,J(t',x'),
\\
&\quad
\text{with } J\in\mathcal{C}.
\end{aligned}
\label{eq:intro_inverse_green}
\end{equation}
Here $\mathcal{C}$ encodes admissible patch sites, Transformer-field directions, sparsity, or depth constraints. Solving this constrained response equation identifies the support and direction of the source $J$, hence the depth-token sites and Transformer-field directions that realize the target effect. Thus the Green function is not only a descriptive propagator; it is the operator that makes patching an inverse reconstruction problem. As in inverse scattering or crystallographic reconstruction, one probes an unknown internal structure, observes an output pattern, and reconstructs the hidden configuration through the response operator. Here the probe is a query or Transformer-field source, the measured pattern is a logit or Transformer-field observable, and the inverse problem seeks to infer a query-conditioned source structure over Transformer-field layer-token sites.

This paper establishes the linear-response foundation of the framework in an empirical Transformer system. We formulate Transformer Field Theory, derive the associated sensitivity fields and Green functions, and test finite-dimensional empirical counterparts of these response objects in GPT-2-style autoregressive Transformers using Transformer-field interventions, Transformer-field differences, and logit-difference observables. The measurements include local linearity and superposition, sensitivity-field predictions, Transformer-field responses, approximate depth composition, high-sensitivity site structure, sliced Green operators, and prompt-induced Transformer-field displacements.

\subsection{Contributions}

The contribution is a field-theoretic response formulation of Transformer-field patching, together with empirical tests of its forward response objects in finite GPT-2-style models. The formalism defines localized source insertions, sensitivity fields, propagated Transformer-field responses, and Green functions over depth and token position. Empirically, we test the corresponding linear-response quantities: local perturbations exhibit a bounded local linear regime; first-order sensitivities predict patch effects across layer-token sites; localized interventions generate structured anisotropic propagation; high-sensitivity sites and sliced Green operators give tractable response descriptions; and prompt-induced Transformer-field displacements partially transfer answer behavior. The same formalism also defines two theoretical targets that motivate the framework. First, patch-site inference asks for sources $J \in \mathcal{C}$ whose propagated responses realize a desired behavioral or Transformer-field shift. Second, model scaling asks whether reduced response information in a smaller model can define candidate patch-site maps in a larger related model under a shared-latent response hypothesis. The continuum notation supplies the organizing limit, while the empirical measurements are discrete response quantities on the depth-token lattice of a finite Transformer.

This viewpoint yields several concrete advantages over purely enumerative patching:
\begin{enumerate}
\item \textbf{Response-based prediction.}
Rather than measuring a separate patching effect for every layer, token, direction, and subspace, one estimates response quantities that predict families of interventions through Eqs.~\eqref{eq:intro_response}--\eqref{eq:intro_Gdef}. This reframes patching from enumerating interventions to estimating sensitivities and response kernels, and gives a field-theoretic reading of gradient-based attribution patching.\cite{nanda2023attribution,syed2023attribution,kramar2024atp}

\item \textbf{Transport and compositionality.}
The propagator describes how localized perturbations move through depth and across tokens. In the ideal linearized theory, propagation from $t_0$ to $t_2$ factors through an intermediate depth $t_1$,
\begin{equation}
G(t_2;\,t_0)
\;\approx\;
G(t_2;\,t_1)\,G(t_1;\,t_0),
\label{eq:intro_semigroup}
\end{equation}
giving a falsifiable signature of approximate dynamical transport.

\item \textbf{Controlled linearity.}
Linear response predicts superposition of small interventions,
\[
\delta y[J_1+J_2]
\approx
\delta y[J_1]+\delta y[J_2].
\]
The failure of this relation identifies the boundary of the perturbative regime and quantifies when patching leaves the local linear approximation.

\item \textbf{Structured response descriptions.}
The field-theoretic formulation organizes patch effects over layer/depth, token position, and Transformer-field dimensions.\cite{cunningham2023sae,gao2024saescaling} In this paper, this structure is instantiated through high-sensitivity site selection and sliced Green operators, giving tractable descriptions of otherwise high-dimensional Transformer-field responses.

\item \textbf{A proposed patch-inference target.}
The adjoint formulation turns patching from measurement into inference: the question ``which patch produces this behavior?'' becomes a constrained variational problem over sources $J$ (Sec.~\ref{sec:action}). The Green function supplies the forward response operator, while sparsity, depth, token, and direction constraints define the admissible patch family.

\item \textbf{A proposed scaling hypothesis.}
The same response geometry suggests a route to model-scale transfer: response fingerprints measured in a smaller model may define candidate patch-site maps in a larger related model, provided the shared-latent response hypothesis is anchored by large-model information such as activations, local probes, or sparse calibration patches (Sec.~\ref{sec:scaling}). The same correspondence can also be used in the reverse direction, mapping large-model response structure back to smaller related models.
\end{enumerate}

The remainder of the paper is organized as follows. Secs.~\ref{sec:continuum}--\ref{sec:response} develop the Transformer field formulation, represent patching as localized source insertion, and derive the associated sensitivity fields and Green functions. Sec.~\ref{sec:action} formulates patch-site inference as an adjoint inverse problem, and Sec.~\ref{sec:scaling} introduces the response-geometry scaling hypothesis. Sec.~\ref{sec:setup} defines the empirical intervention protocol and observables. Secs.~\ref{sec:e1}--\ref{sec:e7} then test the forward response objects in GPT-2-style models, moving from local linearity and sensitivity prediction to Transformer-field propagation, sliced Green operators, and prompt-induced Transformer-field displacements. Sec.~\ref{sec:discussion} summarizes the implications and limitations of the framework.

\subsection{Related Work}

\textbf{Transformers and mechanistic circuits.}
Transformers introduced attention-based sequence modeling and now form the backbone of modern LLMs.\cite{vaswani2017attention}
Mechanistic interpretability seeks to explain model behavior through internal circuits: structured pathways of attention heads, multilayer perceptron (MLP) components, residual features, and token positions.\cite{olah2020circuits,elhage2021framework}
Case studies such as indirect object identification in GPT-2 show that specific behaviors can be localized to multi-layer, multi-token computational pathways.\cite{wang2023ioi}

\textbf{Patching, causal tracing, and attribution.}
Activation patching, causal tracing, and path patching test causal relevance by replacing internal activations or isolating paths and measuring the resulting change in output.\cite{elhage2021framework,wang2023ioi,goldowskyDill2023pathpatching,zhang2023patching}
These methods reveal where behaviorally meaningful information is carried. However, applying them exhaustively can be expensive: the number of interventions grows with the number of layers, tokens, components, directions, and subspaces tested.
Attribution patching replaces measured interventions with a first-order proxy: the patch effect is approximated by the inner product between an activation displacement and the gradient of the observable.\cite{nanda2023attribution,syed2023attribution,kramar2024atp}
In our notation, this is the discrete version of Eq.~\eqref{eq:intro_response}. Our first-order response experiments measure the regime in which this prediction is accurate for Transformer-field interventions, including settings where linear proxies can be delicate.\cite{nanda2023attribution,kramar2024atp}
Causal scrubbing provides a complementary intervention-based methodology for testing mechanistic hypotheses against model behavior.\cite{chan2022causal}

\textbf{Continuous-depth dynamics and adjoints.}
Residual networks admit a continuous-depth interpretation in which layerwise updates approximate differential equations.\cite{he2015resnet,chen2018neuralode}
This viewpoint treats representations as trajectories and makes sensitivities, propagators, and adjoint equations natural objects.
Training continuous-depth networks can be formulated through optimal-control and adjoint-state ideas: a backward costate equation gives the sensitivity of a terminal objective to the state trajectory.\cite{chen2018neuralode,pontryagin1962}
We use the same structure for interpretability: the adjoint becomes the sensitivity field, and patch selection becomes a constrained source-inference problem.

\textbf{Response-field theory.}
Linear response theory predicts the effect of localized perturbations through sensitivities, propagators, and Green functions.\cite{kubo1957}
For dynamical systems, the Martin--Siggia--Rose--Janssen--De~Dominicis (MSRJD) formalism represents dynamics by an action with an auxiliary response field whose insertions generate response functions.\cite{MSR1973,janssen1976,dedominicis1976}
We use this language to treat Transformer-field patching as localized source insertion and mechanistic influence as empirical response propagation over depth and token position.

\textbf{Prompt transformations, task vectors, and steering directions.}
A parallel line of work shows that behaviorally meaningful transformations can be represented as additive directions in the activation space of the Transformer field: task vectors summarize in-context demonstrations,\cite{hendel2023taskvectors} function vectors encode task-relevant transformations in autoregressive Transformer activations,\cite{todd2024functionvectors} and activation addition or representation engineering steers behavior by adding contrastive directions at inference.\cite{turner2023activation,zou2023representation}
Our prompt-displacement experiments place these activation-space directions into the same field-theoretic framework: a prompt difference becomes a localized source, and its effect is measured through the Transformer-field response it induces.

\medskip
Taken together, these lines of work provide causal probes, activation directions, and continuous-depth tools, but they do not by themselves formulate Transformer patching as an inverse Green-function problem. Our contribution is to build that formulation for the Transformer field: patching becomes localized source insertion, attribution becomes a sensitivity-field prediction, propagation is measured through empirical Green-operator responses, and patch selection becomes an adjoint inverse problem. The point is not only to measure which activations matter, but to make internal mechanism reconstruction mathematically well posed. From query-induced responses and output observables, the Green-function equation supplies the forward operator for an inverse problem whose solution would identify patch sites, Transformer-field directions, and query-conditioned graph-like response structure.

\section{Depth--Token Continuum: Residual Stream as a Transformer Field}
\label{sec:continuum}

\subsection{Discrete-to-continuous mapping}

Let $\ell\in\{0,\dots,L\}$ index Transformer layers and $j\in\{1,\dots,n\}$ index tokens. We introduce:
\begin{itemize}
\item Depth-time: $t\in[0,T]$ with step $\Delta t=T/L$ and $t_\ell=\ell\Delta t$.
\item Token-space: $x$ as the coordinate associated with token index. At minimum, $x$ is a 1D lattice site ($x=j$); in long-context limits it may be idealized as a continuum coordinate.
\end{itemize}
As described in the introduction, Eq.~\eqref{eq:intro_discrete_transformer_field}
defines the discrete Transformer field \(\rfield_\ell(x)\), while
Eq.~\eqref{eq:intro_continuum_transformer_field} defines its continuum
idealization \(\rfield(t,x)\).

\subsection{Basic continuum object: $\rfield(t,x)$}

Across Transformer blocks, the Transformer field can be written schematically as the update
\begin{align}
\rfield_{\ell+1}(x)
&=
\rfield_{\ell}(x)
+
\Delta t\,
\Fop_{\ell}[\rfield_{\ell}](x),
\nonumber\\
&\qquad
\Delta t=\frac{T}{L},
\label{eq:discrete_transformer_field_update}
\end{align}
where $\Fop_{\ell}$ denotes the effective Transformer-block update, including attention, MLP, normalization, and layer-specific parameters.

The formal limit $L\to\infty$ gives the depth evolution
\begin{equation}
\partial_t \rfield(t,x)
=
\Fop_t[\rfield](x),
\label{eq:continuous_dynamics}
\end{equation}
where $\Fop_t$ is a depth-dependent vector field governing the evolution of the Transformer field. In this limit, layer-specific parameters are represented as depth-dependent coefficients or operators.\cite{he2015resnet,chen2018neuralode}

The structural feature we use is that the attention contribution to $\Fop_t$ is nonlocal over token space: the update at $x$ depends on $\rfield(t,y)$ across tokens $y$. Consequently, the linearized response couples token sites, and localized perturbations can spread across token position. We do not require an explicit attention kernel below, and treat $\Fop_t$ as a generic nonlocal Transformer-field operator.

\section{Patching as a Localized Source Insertion}
\label{sec:patching}

\subsection{Patching in field language}

\begin{figure}[t]
\centering
\includegraphics[width=\linewidth]{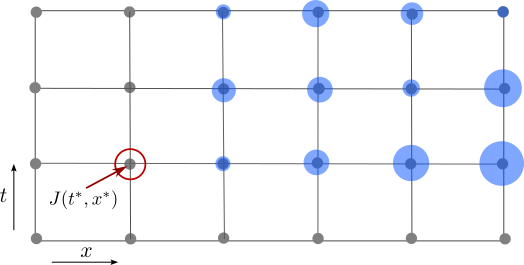}
\caption{\textbf{Patching as a localized source insertion.} A patch at $(t^{\ast},x^{\ast})$ is modeled as a source term $J(t,x)$ that injects a localized perturbation into the depth evolution. Attention then spreads its influence nonlocally across token positions.}
\label{fig:patch_defect}
\end{figure}

Activation patching replaces an internal activation at layer $\ell^{\ast}$ and token $x^{\ast}$ with a value taken from a source run.\cite{elhage2021framework,wang2023ioi,zhang2023patching} In the depth-token continuum, this becomes a localized intervention at $(t^{\ast},x^{\ast})$. Since \(t\) denotes network depth, \(\partial_t\rfield\) describes how the Transformer field evolves from layer to layer; the patch appears as an added localized forcing term:
\begin{equation}
\partial_t \rfield(t,x)
=
\Fop_t[\rfield](x)
+
J(t,x),
\label{eq:patched_dynamics}
\end{equation}
where an instantaneous patch is represented by
\begin{equation}
\begin{aligned}
J(t,x)
&=
\delta(t-t^{\ast})\,\delta(x-x^{\ast})
\\
&\quad\times
P\Big(
\rfield_{\mathrm{src}}(t^{\ast},x^{\ast})
-
\rfield^{0}(t^{\ast},x^{\ast})
\Big).
\end{aligned}
\label{eq:J_instant}
\end{equation}
Here \(\rfield_{\mathrm{src}}\) is the source-run Transformer field and \(\rfield^{0}\) is the unpatched target-run field. The delta factors localize the source in depth and token position, while \(P\) projects onto the patched subspace: an attention-head output subspace, an MLP direction, a learned feature direction, or the full Transformer-field vector \((P=\mathrm{Id})\). Figure~\ref{fig:patch_defect} summarizes this representation: the patch is a localized source insertion in the depth evolution, and attention spreads its influence across token positions.

\subsection{Patching over an interval of layers}

When the intervention replaces activations over a finite depth interval rather than at a single depth, the depth impulse is replaced by an interval-supported source:
\begin{equation}
\begin{aligned}
J(t,x)
&=
\chi_{[t_1,t_2]}(t)\,\delta(x-x^{\ast})
\\
&\quad\times
P\Big(
\rfield_{\mathrm{src}}(t,x^{\ast})
-
\rfield^{0}(t,x^{\ast})
\Big).
\end{aligned}
\label{eq:J_interval}
\end{equation}
Here \(\chi_{[t_1,t_2]}(t)\) is the indicator, or characteristic, function of the interval \([t_1,t_2]\): it equals one for \(t\in[t_1,t_2]\) and zero otherwise. Thus Eq.~\eqref{eq:J_instant} represents an instantaneous localized patch, while Eq.~\eqref{eq:J_interval} represents the continuum analogue of a discrete patching window across layers.\cite{goldowskyDill2023pathpatching}

\section{Mechanistic Pathways as Response Functions}
\label{sec:response}

\subsection{Response kernels and functional derivatives}

The patched field equation describes how a Transformer-field perturbation enters the depth evolution. To identify which perturbations affect an output, we study the response of an observable $y$ to infinitesimal changes of the Transformer field,
\begin{equation}
\frac{\delta y}{\delta \rfield(t^{\ast},x^{\ast})}.
\end{equation}
Component indices are suppressed. The corresponding two-point response, or Green function, is
\begin{equation}
G(t,x; t^{\ast},x^{\ast})
:=
\frac{\delta \rfield(t,x)}
     {\delta \rfield(t^{\ast},x^{\ast})}.
\label{eq:G_def}
\end{equation}
It is the continuum analogue of the map from a patch site $(\ell^{\ast},x^{\ast})$ to its downstream effect across later layers and tokens. In this sense, \(G\) is the infinitesimal mechanistic pathway kernel: it records how a local Transformer-field perturbation is transported by the model's depth-token dynamics.

\subsection{Linear response equation for $G$}

The Green function is obtained by linearizing the nonlinear Transformer-field evolution around an unpatched trajectory $\rfield^{0}(t,x)$. Write
\[
\rfield(t,x)
=
\rfield^{0}(t,x)
+
\delta\rfield(t,x),
\]
and keep only first-order terms in $\delta\rfield$. For the patched dynamics with external source $J(t,x)$, the resulting first-order perturbation obeys
\begin{equation}
\partial_t \delta \rfield(t,x)
=
\int \diff y\; \Lop_t(x,y)\,\delta \rfield(t,y)
+
J(t,x),
\label{eq:linearized_dynamics}
\end{equation}
where
\[
\Lop_t(x,y)
:=
\left.
\frac{\delta \Fop_t[\rfield](x)}
     {\delta \rfield(t,y)}
\right|_{\rfield=\rfield^{0}}
\]
is the Fr\'echet derivative of the Transformer-field evolution operator along the clean trajectory. Because attention is nonlocal in token position, $\Lop_t(x,y)$ couples distinct token sites.

The Green function is the fundamental solution of this linearized depth equation. Equivalently, it is the response generated by an infinitesimal impulse at $(t^{\ast},y)$ and then propagated forward through the linearized dynamics:
\begin{align}
\partial_t G(t,\cdot; t^{\ast},\cdot)
&=
\Lop_t\,G(t,\cdot; t^{\ast},\cdot),
\qquad t>t^{\ast},
\label{eq:G_eom}
\\
G(t^{\ast},x;t^{\ast},y)
&=
\delta(x-y).
\nonumber
\end{align}
Thus Eq.~\eqref{eq:G_eom} is the continuum version of following the effect of a patch through later Transformer layers. In the empirical sections, we do not estimate the full Green operator; instead, we measure tractable response objects derived from it: Transformer-field responses, approximate composition tests, and sliced Green operators.

\subsection{Patching effect on observables}

Let $y$ be an output observable, such as a next-token logit difference. In linear response, the effect of a patch source $J$ is
\begin{equation}
\begin{aligned}
\delta y
&\approx
\int \diff t\,\diff x\;
a(t,x)\,J(t,x),
\\
a(t,x)
&:=
\frac{\delta y}{\delta \rfield(t,x)} .
\end{aligned}
\label{eq:dy_response}
\end{equation}
The sensitivity field $a(t,x)$ is the readout-side response: it tells how an infinitesimal Transformer-field perturbation at $(t,x)$ changes the observable. Equivalently, a localized patch first propagates forward through the Green function and is then read out by the final-depth sensitivity. Thus \(a\) and \(G\) are complementary response objects: \(G\) describes transport through the Transformer-field dynamics, while \(a\) describes how transported perturbations affect the observable.

\section{Action Principle and the Patch-Inference Problem}
\label{sec:action}

The response objects above answer the forward question: what does a given patch do? The inverse question is a mechanistic localization problem: which source, placed where in the Transformer field, produces a prescribed behavioral effect? We call this the \emph{patch-inference problem}. It is a constrained variational problem over source terms $J$: the source must obey admissibility constraints, enter the Transformer-field dynamics, and induce a target observable or Transformer-field displacement. In the linear regime, the Green function is the kernel of the corresponding inverse integral equation. As a useful metaphor, this is analogous to inverse scattering or crystallographic reconstruction: a measured output pattern is related to a hidden structure through a forward response operator. Here the query plays the role of the probe that fixes the computational context, and the measured pattern is a logit or Transformer-field observable. The hidden object is the query-conditioned causal response structure of the Transformer system, represented by the support and Transformer-field directions of the source $J$.

\subsection{Action and adjoint field}

The constrained formulation is encoded by an action. We enforce the Transformer-field dynamics of Eq.~\eqref{eq:continuous_dynamics} with an adjoint field $\adj(t,x)$ and add a terminal readout:
\begin{align}
S[\rfield,\adj]
&=
\int_0^T \!\diff t \int \!\diff x\;
\adj(t,x)
\Big(
\partial_t \rfield(t,x)-\Fop_t[\rfield](x)
\Big)
\nonumber\\
&\quad
+
S_{\mathrm{readout}}\big[\rfield(T,\cdot)\big].
\label{eq:action}
\end{align}
The adjoint field is a continuum Lagrange multiplier: it enforces the dynamics at every depth-token point and records how variations of the Transformer-field trajectory affect the terminal readout. Stationarity, $\delta S=0$, gives the Euler--Lagrange optimality equations. Variation with respect to $\adj$ recovers the forward Transformer-field dynamics; variation with respect to $\rfield$ gives the backward adjoint equation
\begin{align}
-\,\partial_t \adj(t,x)
&=
\int \diff y\; \Lop_t^{\dagger}(y,x)\,\adj(t,y),
\label{eq:adjoint_eom}
\\
\adj(T,x)
&=
\frac{\delta S_{\mathrm{readout}}}{\delta \rfield(T,x)} .
\nonumber
\end{align}
Here $\Lop_t^{\dagger}$ is the adjoint of the linearized operator in Eq.~\eqref{eq:linearized_dynamics}. In optimal-control terminology, Eq.~\eqref{eq:adjoint_eom} is the costate equation; it is the standard adjoint-state construction used for continuous-depth networks.\cite{chen2018neuralode,pontryagin1962} Solving this equation backward from the terminal readout condition gives the sensitivity field:
\begin{equation}
\adj(t,x)
=
\frac{\delta y}{\delta \rfield(t,x)}
=
a(t,x).
\label{eq:lambda_is_a}
\end{equation}
Thus the forward Green function $G$ from Sec.~\ref{sec:response} is the forward response operator, while $\adj=a$ is the backward sensitivity. They are the forward and adjoint faces of the same linear-response structure.

\subsection{Patches as operator insertions}

A patch source modifies the Transformer-field dynamics by adding \(J\). Using generating-functional notation, the patch is represented by the exponentiated insertion
\begin{equation}
\exp\left(
\int \diff t \int \diff x\, \lambda(t,x)J(t,x)
\right).
\label{eq:source_insertion}
\end{equation}

This insertion turns the action formalism into a generating object for response: functional derivatives with respect to $J$ produce the response of the readout and of the Transformer-field trajectory to localized sources. The first functional derivative gives the response formula in Eq.~\eqref{eq:dy_response} with $a=\adj$; second functional derivatives give Green functions. This is the deterministic analogue, for Transformer-field dynamics, of the MSRJD response-field construction.\cite{MSR1973,janssen1976,dedominicis1976} In this formulation, patching is not an external operation added after the fact; it is represented inside the variational structure as a source coupled to the field that measures response.

\subsection{The inference problem}

Patch inference asks for a source $J$ that realizes a prescribed observable shift $\Delta y^{\star}$ at minimal cost, while satisfying both the behavioral constraint and the patched Transformer-field dynamics:
\begin{equation}
\begin{aligned}
\min_{J}\quad
& C[J]
\\[-1mm]
\text{s.t.}\quad
& \delta y[J]=\Delta y^{\star}
\\
& \text{and}\quad
\partial_t \rfield
=
\Fop_t[\rfield]+J .
\end{aligned}
\label{eq:inverse_problem}
\end{equation}
The cost $C[J]$ defines the admissible patch family: energy, sparsity, depth support, token support, direction constraints, or combinations of these. Introducing multipliers gives the corresponding optimality system: forward dynamics for $\rfield$, backward adjoint dynamics for $\adj$, and a stationarity condition selecting the admissible source. Thus patch localization is recast as an inverse problem constrained by the Transformer-field dynamics, rather than as a search over isolated interventions.

In the local linear regime, the inverse problem becomes a constrained response equation. For a prescribed Transformer-field target, the admissible source satisfies
\begin{equation}
\begin{aligned}
\Delta\rfield_{\mathrm{target}}(t,x)
&\approx
\int \diff t'\,\diff x'\;
G(t,x;t',x')\,J(t',x'),
\\
&\quad
\text{with } J\in\mathcal{C}.
\end{aligned}
\label{eq:inverse_green_action}
\end{equation}
Here $\mathcal{C}$ specifies the allowed patch sites, directions, and sparsity pattern. The Green function is the kernel of this inverse integral equation. Solving the equation identifies the support and direction of $J$, namely the depth-token locations and Transformer-field directions whose propagated response realizes the target. The forward measurements in this paper establish the linear-response objects needed to pose this reconstruction problem in finite Transformer models: sensitivities, Transformer-field responses, high-sensitivity structures, and sliced Green operators.

\section{Model Scaling of Field-Theoretic Response}
\label{sec:scaling}

A motivation for the field-theoretic formulation is that response structure need not be tied to one particular model discretization. If models in the same family approximate a common latent response geometry, then response information measured in a smaller model $M$ can suggest candidate patch sites in a larger model $M'$ without estimating the full large-model Green operator. By response geometry, we mean the geometry induced by response fingerprints: sites are compared by what affects them and what they affect, rather than only by layer index, component label, or activation similarity.

Here we propose a scaling hypothesis: reduced response information in a smaller model $M$, together with local anchors in a larger model $M'$, can define a candidate map between patch-relevant sites across scale.

\subsection{Models as discretizations of a latent response operator}

Let normalized depth be $s=t/T\in[0,1]$, and write a discrete patch site as
\[
a=(\ell,x,i),
\]
where $\ell$ is the layer, $x$ is the token position, and $i$ indexes the Transformer-field component, with $1\le i\le d_{\mathrm{model}}^{(M)}$ for model $M$. In the continuum notation, this site is mapped to $(s,x,u)$, where $u$ is the latent counterpart of the model-specific Transformer-field component index $i$. The proposed scaling picture treats $M$ and $M'$ as two model-specific discretizations of a shared latent response operator, conditioned on a query $q$,
\[
\mathcal{G}(q;\,s,x,u;\,s',x',u').
\]
Their Green operators are then finite model-specific views of this latent operator:
\begin{equation}
\begin{aligned}
G^{(M)}(q)
&\approx
\Pi_M\,\mathcal{G}(q)\,\Pi_M^{\ast},
\\
G^{(M')}(q')
&\approx
\Pi_{M'}\,\mathcal{G}(q')\,\Pi_{M'}^{\ast}.
\end{aligned}
\label{eq:scale_compression}
\end{equation}
The maps $\Pi_M$ and $\Pi_{M'}$ encode how each finite model samples or projects the latent response geometry. Concretely, they select the depth grid, token grid, and Transformer-field component basis of each model, and may also account for differences in width or Transformer-field component coordinates. Thus $G^{(M)}$ and $G^{(M')}$ are not assumed to be identical Green operators; they are different finite coordinate representations of a shared latent response structure.

This shared-latent response hypothesis is the field-theoretic analogue of cross-model representational convergence and model stitching.\cite{huh2024platonic,lenc2015stitching,bansal2021stitching} Under this hypothesis, the scaling problem is to estimate a discretization map
\[
T_{M\to M'}=(T_{\mathrm{layer}},T_{\mathrm{token}},T_{\mathrm{comp}})
\]
that carries patch-relevant sites in $M$ to corresponding candidate sites in $M'$.

\subsection{Response fingerprints and the intertwining condition}

A site is characterized by its incoming and outgoing response profile,
\begin{equation}
\phi(a)
:=
\big(G[:,a],\,G[a,:]\big),
\label{eq:scale_fingerprint}
\end{equation}
namely what affects $a$ and what $a$ affects. Given a reduced decomposition
\[
G\approx U_r\Sigma_r V_r^{\top},
\]
we define the compact response fingerprint
\[
\tilde{\phi}(a)
=
\big(U_r(a,:)\Sigma_r,\;V_r(a,:)\Sigma_r\big).
\]
The scaling hypothesis says that corresponding sites across model size should have compatible response fingerprints. Equivalently, at the level of linearized response dynamics, one seeks an approximate intertwiner $\Psi$ satisfying
\begin{equation}
\Lop^{(M')}_s \Psi
\approx
\Psi\Lop^{(M)}_s
\quad\Longrightarrow\quad
G^{(M')}\Psi
\approx
\Psi G^{(M)} .
\label{eq:scale_intertwine}
\end{equation}
Thus an intertwiner of the generators would align Green-operator structure across models. The matrix $\Psi$ should be understood as a soft correspondence between sites of $M$ and sites of $M'$. A discrete patch-site map is then obtained from its columns:
\begin{equation}
T_{M\to M'}(a)
=
\arg\max_{a'} |\Psi_{a',a}| .
\label{eq:scale_transfer}
\end{equation}

\subsection{Estimation without the full large-model Green operator}

The point of the scaling hypothesis is that the correspondence can be estimated without forming the full large-model Green operator. One possible formulation is
\begin{equation}
\Psi^{\star}
=
\arg\min_{\Psi}\;
\big\|
\Lop^{(M')}_s\Psi
-
\Psi\Lop^{(M)}_s
\big\|^2
+
\lambda\,\mathcal{R}(\Psi),
\label{eq:scale_estimation}
\end{equation}
using local Jacobian--vector products or sparse response probes in $M'$. Alternatively, reduced response fingerprints may be matched by a geometry-preserving optimal-transport objective under a depth-locality constraint
\[
|s(a)-s(a')|<\varepsilon .
\]
where $s(a)$ denotes the normalized depth coordinate of site $a$. This would give a response-geometry-preserving map rather than a heuristic layer alignment.\cite{memoli2011gromov,peyre2016gromov}

Identifiability requires more than $G^{(M)}$ alone. The shared-latent response hypothesis must be supplemented by large-model anchors such as weights, activations, local linear probes, or a small set of calibration patches. With such anchors, the full large-model Green operator is replaced by three tractable ingredients: response geometry from the small model, local descriptors in the larger model, and sparse validation by direct patching. The proposed scaling workflow is therefore to measure $G^{(M)}$ or its reduced response geometry, predict candidate sites in $M'$, and validate the top transferred sites by sparse patching. In this way, the field-theoretic formulation suggests a route from patch-site inference in one model to response scaling across a model family.

\section{Empirical Setup and Operational Recipe}
\label{sec:setup}

\subsection{Operational response quantities}

The empirical implementation uses the discrete Transformer-field
$\rfield_\ell(x)\in\mathbb{R}^{d_{\mathrm{model}}}$ rather than the formal continuum.
The observable is a scalar logit difference between candidate tokens at the autoregressive readout,
\begin{equation}
y
=
\mathrm{logit}(w_{\mathrm{target}})
-
\mathrm{logit}(w_{\mathrm{ref}}),
\label{eq:observable}
\end{equation}
measured on a fixed prompt or controlled prompt pair.

A local Transformer-field intervention at layer $\ell^{\ast}$ and token $x^{\ast}$ is
\begin{equation}
\rfield_{\ell^{\ast}}(x^{\ast})
\mapsto
\rfield_{\ell^{\ast}}(x^{\ast})+\epsilon J,
\qquad
J\in\mathbb{R}^{d_{\mathrm{model}}}.
\label{eq:discrete_patch}
\end{equation}
Here $\rfield_{\ell^{\ast}}(x^{\ast})$ is the Transformer-field vector at token $x^{\ast}$ after block $\ell^{\ast}$, and all predictions and measurements use this same Transformer-field site. The measured scalar effect is
\begin{equation}
\delta y_{\mathrm{meas}}(\epsilon,J)
=
y(\rfield+\epsilon J)-y(\rfield),
\label{eq:dy_meas}
\end{equation}
and the first-order response prediction is
\begin{equation}
\begin{aligned}
\delta y_{\mathrm{pred}}(\epsilon,J)
&=
\epsilon\,a(\ell^{\ast},x^{\ast})\cdot J,
\\
a(\ell,x)
&:=
\frac{\partial y}{\partial \rfield_\ell(x)} .
\end{aligned}
\label{eq:dy_pred}
\end{equation}

For propagation experiments, the measured object is the Transformer-field response
\begin{equation}
\delta\rfield_\ell(x)
:=
\rfield_\ell^{\mathrm{patched}}(x)
-
\rfield_\ell^{0}(x),
\qquad
\ell\ge \ell^{\ast}.
\label{eq:dR}
\end{equation}

\subsection{Prompt families and experimental details}

The empirical experiments use capital-completion prompts. The main fixed prompt is a newline-separated list of country--capital facts ending with ``The capital of Spain is'', with the observable usually taken as the logit difference between the answer tokens \mbox{`` Madrid''} and \mbox{`` London''}. The Transformer-field sweeps compare this nonrepetitive list prompt with a repetitive capital-fact variant, while the sensitivity-concentration experiment uses a prose version of the same European-capital context. The prompt-displacement experiments use matched same-template pairs such as ``The capital of Spain is'' and ``The capital of France is'', with answer tokens such as \mbox{`` Madrid''} and \mbox{`` Paris''}.

The empirical sections then move from local linearity and scalar sensitivity prediction to Transformer-field propagation, approximate composition, reduced response descriptions, sliced Green operators, and prompt-induced Transformer-field displacements. Together these experiments instantiate the forward response objects introduced above on the finite depth--token lattices of GPT-2-style Transformer models.

\begin{figure*}[t]
\centering
\includegraphics[width=\textwidth]{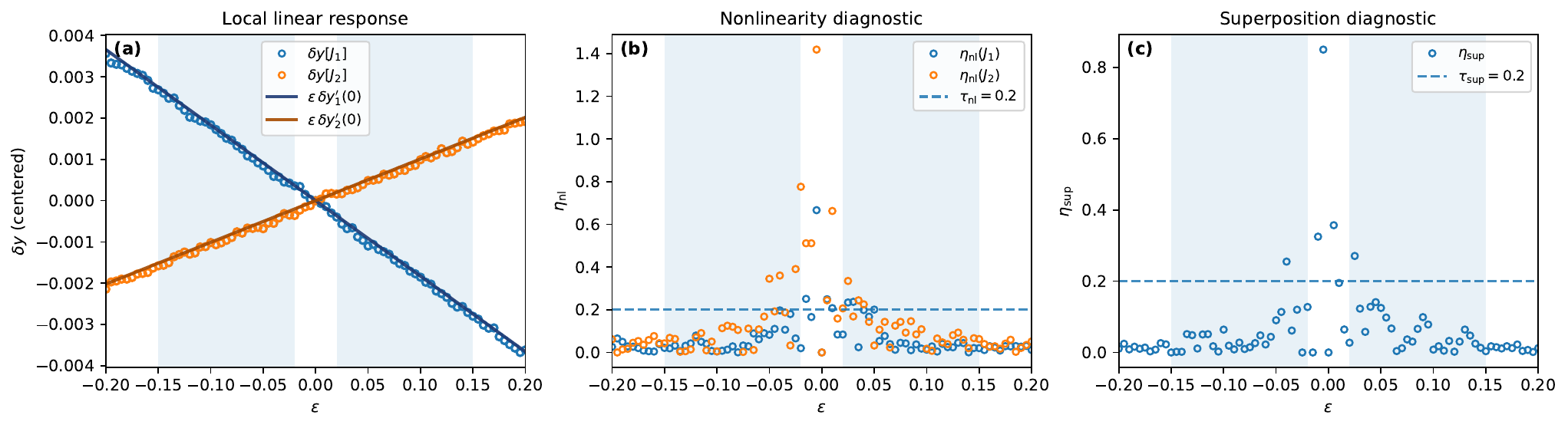}
\caption{\textbf{Local linearity and superposition diagnostics.} At a fixed layer-token site with $\ell^\ast=7$ and $x^\ast$ the final token in the prompt, two directions $J_1,J_2$ and their sum are injected over an amplitude sweep. Panel (a) compares measured centered output changes $\delta y(\epsilon)$ with linear fits estimated near $\epsilon=0$. Panel (b) shows the relative nonlinearity $\eta_{\mathrm{nl}}$, and panel (c) shows the superposition error $\eta_{\mathrm{sup}}$. The horizontal reference lines mark a $0.2$ visual reference level. The shaded region marks the $\epsilon$ band used for slope estimation.}
\label{fig:e1_linearity}
\end{figure*}

\section{Local Linearity and Superposition}
\label{sec:e1}

A response-theoretic description requires a local linear regime. For a fixed Transformer-field site $(\ell^{\ast},x^{\ast})$ and directions $J,J_1,J_2$, we sweep the perturbation amplitude $\epsilon$ and measure the scalar response $\delta y$. Linear response predicts
\begin{equation}
\delta y[\epsilon J]
\approx
\epsilon\,\chi_J,
\qquad
\chi_J
:=
a(\ell^{\ast},x^{\ast})\cdot J,
\label{eq:e1_linear}
\end{equation}
and superposition of small interventions,
\begin{equation}
\delta y[\epsilon(J_1+J_2)]
\approx
\delta y[\epsilon J_1]
+
\delta y[\epsilon J_2].
\label{eq:e1_superposition}
\end{equation}

We measure nonlinearity by
\begin{equation}
\eta_{\mathrm{nl}}(\epsilon)
:=
\frac{
\left|\delta y(\epsilon)-\epsilon\,\delta y'(0)\right|
}{
\left|\epsilon\,\delta y'(0)\right|
},
\label{eq:e1_eta_nl}
\end{equation}
and superposition error by
\begin{equation}
\eta_{\mathrm{sup}}(\epsilon)
:=
\frac{
\left|
\delta y[\epsilon(J_1+J_2)]
-
\delta y[\epsilon J_1]
-
\delta y[\epsilon J_2]
\right|
}{
\max\left(
|\delta y[\epsilon J_1]|+|\delta y[\epsilon J_2]|,
\varepsilon_0
\right)
}.
\label{eq:e1_eta_sup}
\end{equation}
The slope $\delta y'(0)$ is estimated from a near-zero band, excluding numerically unstable denominators at $\epsilon\approx0$. Reference levels for $\eta_{\mathrm{nl}}$ and $\eta_{\mathrm{sup}}$ help visualize the perturbative band. In the superposition test, $J_1+J_2$ is injected without renormalization; for unit-normalized, near-orthogonal directions, $J_1+J_2$ has norm approximately $\sqrt{2}$, so the matched-$\epsilon$ superposition sweep also probes a larger total perturbation norm.

The empirical result is a bounded local linear regime for Transformer-field interventions. Within this band, scalar responses approximately follow a slope through the origin and superposition error remains controlled; beyond it, nonlinear and interaction effects increase. This regime provides a local calibration of the first-order response approximation, while later experiments also probe larger finite-amplitude interventions.

As shown in Fig.~\ref{fig:e1_linearity}, $\eta_{\mathrm{nl}}$ and $\eta_{\mathrm{sup}}$ remain mostly below the $0.20$ reference level across the displayed perturbative range, with the main caveat that very small $|\epsilon|$ can produce denominator-driven artifacts.

\section{Sensitivity-Field Prediction of Scalar Responses to Patching}
\label{sec:e2}

We next test whether the local sensitivity field predicts measured scalar responses to patching. For each Transformer-field site $(\ell,x)$, define
\begin{equation}
a(\ell,x)
=
\frac{\partial y}{\partial \rfield_\ell(x)}
\in
\mathbb{R}^{d_{\mathrm{model}}},
\label{eq:e2_gradient}
\end{equation}
computed on the clean run. For a Transformer-field direction $J$, linear response predicts
\begin{equation}
\delta y_{\mathrm{pred}}
=
\epsilon\,a(\ell,x)\cdot J .
\label{eq:e2_prediction}
\end{equation}
We compare this prediction with the measured scalar response $\delta y_{\mathrm{meas}}$ obtained by injecting $\epsilon J$ at the same Transformer-field site. Equations~\eqref{eq:e2_gradient}--\eqref{eq:e2_prediction} are the Transformer-field form of attribution patching: a gradient--activation inner product replaces a direction-by-direction intervention sweep.\cite{nanda2023attribution,syed2023attribution,kramar2024atp}

\begin{figure}[t]
\centering
\includegraphics[width=\columnwidth]{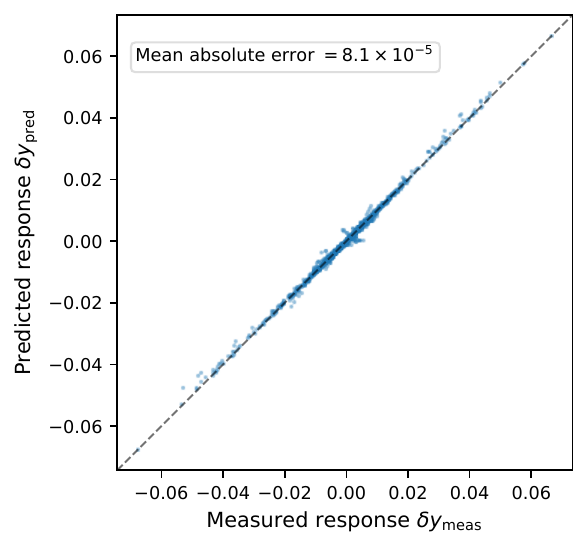}
\caption{\textbf{Sensitivity-field prediction of measured scalar responses.}
Each point compares the measured logit-difference response $\delta y_{\mathrm{meas}}$ from injecting $\epsilon J$ at a Transformer-field site with the first-order prediction $\delta y_{\mathrm{pred}}=\epsilon\,a(\ell,x)\cdot J$, with near-zero amplitudes $\epsilon$ excluded from the plotted sample. The horizontal axis is $\delta y_{\mathrm{meas}}$, the vertical axis is $\delta y_{\mathrm{pred}}$, and the dashed line marks equality. A single autograd sensitivity field at a site predicts responses for many directions $J$, replacing a direction-by-direction patching sweep.}
\label{fig:e2_prediction}
\end{figure}

\begin{figure*}[t]
\centering
\includegraphics[width=0.95\textwidth]{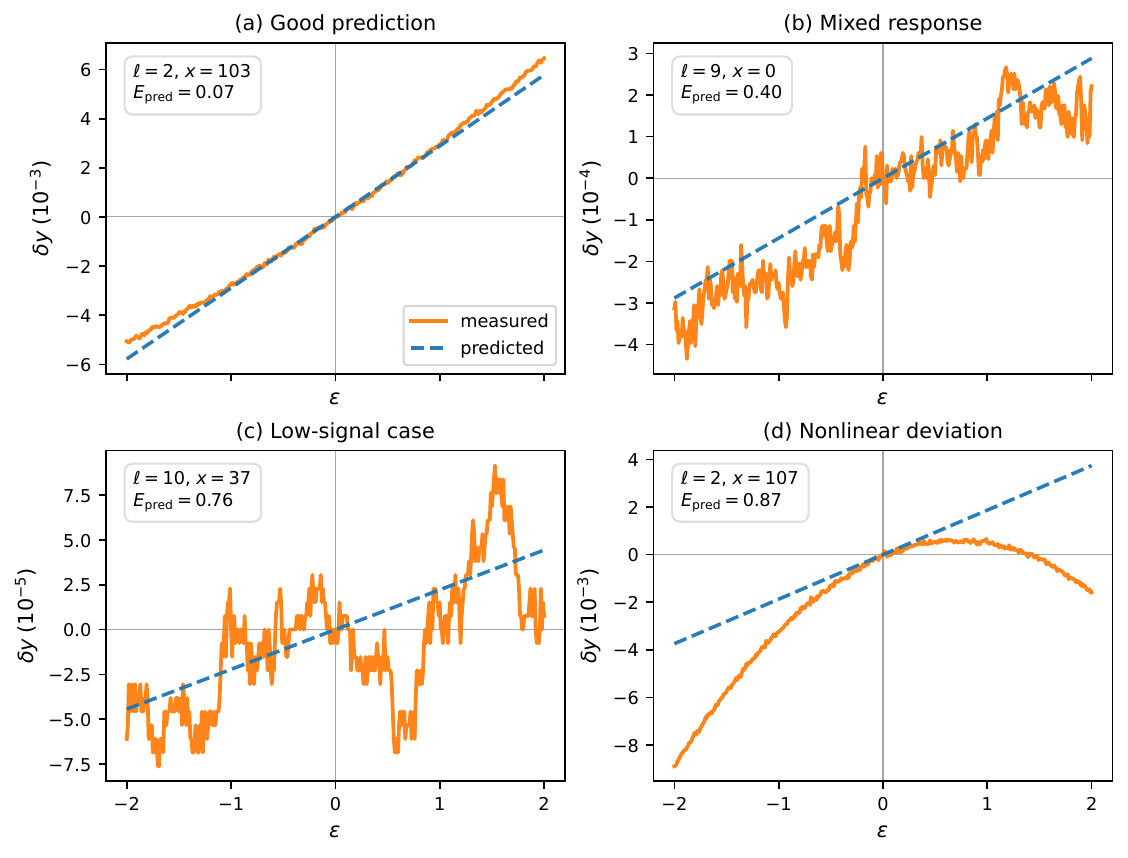}
\caption{\textbf{Sensitivity-field prediction diagnostics.} Representative amplitude sweeps compare the measured scalar response $\delta y(\epsilon)$ with the first-order prediction $\delta y_{\text{pred}}=\epsilon\,a(\ell,x)\cdot J$ at individual layer-token sites. The orange curve is measured, the dashed blue line is predicted, and each in-panel annotation reports the site $(\ell,x)$ together with $E_{\mathrm{pred}}$, a sweep-level analogue of the relative prediction error in Eq.~\eqref{eq:e2_rel_err} formed from $\epsilon$-averages of the absolute prediction error and measured response magnitude. The examples show a good linear prediction, an intermediate mixed-response regime, a low-signal denominator artifact, and a strong nonlinear deviation. These traces explain which site-level behaviors underlie the aggregate measured-versus-predicted comparison in Fig.~\ref{fig:e2_prediction}.}
\label{fig:e2_prediction_diagnostics}
\end{figure*}

Figures~\ref{fig:e2_prediction} and~\ref{fig:e2_prediction_diagnostics} test Eq.~\eqref{eq:e2_prediction} at two resolutions. Figure~\ref{fig:e2_prediction} pools sites, directions, and amplitudes, plotting the measured $\delta y_{\mathrm{meas}}$ against the gradient prediction $\delta y_{\mathrm{pred}}=\epsilon\,a(\ell,x)\cdot J$; alignment with the diagonal is the aggregate evidence that the clean-run gradient predicts finite-patch responses. Figure~\ref{fig:e2_prediction_diagnostics} shows the same prediction before pooling: for selected layer-token sites, $\epsilon$ is swept, the blue line is $\epsilon\,a(\ell,x)\cdot J$, and the orange curve is the measured $\delta y(\epsilon)$. These traces expose the good, mixed, low-signal, and nonlinear cases behind the scatter. Here $\epsilon$ is the scalar amplitude multiplying the Transformer-field direction $J$ in the patch $\rfield_\ell(x)\mapsto\rfield_\ell(x)+\epsilon J$.

Prediction quality is summarized by
\begin{align}
E_{\mathrm{abs}}
&=
\left|
\delta y_{\mathrm{meas}}
-
\delta y_{\mathrm{pred}}
\right|,
\label{eq:e2_abs_err}
\\
E_{\mathrm{rel}}
&=
\frac{
\left|
\delta y_{\mathrm{meas}}
-
\delta y_{\mathrm{pred}}
\right|
}{
|\delta y_{\mathrm{meas}}|+\varepsilon_0
}.
\label{eq:e2_rel_err}
\end{align}
The relative metric is most interpretable when the measured signal is not dominated by the small-denominator regularizer; absolute error remains informative throughout. These error quantities are used to classify the representative sweeps in Fig.~\ref{fig:e2_prediction_diagnostics}, where large relative error can indicate either genuine finite-amplitude nonlinearity or a low-signal denominator artifact. 

The empirical result is that autograd sensitivities predict measured scalar responses in the perturbative regime. Thus $a(\ell,x)$ acts as a predictive compression of patching outcomes: one sensitivity evaluation replaces a sweep over Transformer-field directions, realizing the operational content of Eq.~\eqref{eq:intro_response}.

\begin{algorithm}[!t]
\caption{Downstream Transformer-field response $\delta\rfield$}
\label{alg:field}
\KwIn{Transformer-field evolution; query $q$; source sites $\mathcal{S}$;
      direction $J$, $\|J\|=1$; amplitude $\epsilon$}
\KwOut{mean response $\langle\|\delta\rfield\|\rangle(\Delta\ell,\Delta x)$
       over $\Delta\ell\ge0,\Delta x\ge0$}

Compute the clean Transformer field $\rfield^{0}_{\ell}(x)$ for the query $q$\;

\ForEach{$(\ell^{\ast},x^{\ast})\in\mathcal{S}$}{
  Apply the localized Transformer-field perturbation
  \[
  \rfield_{\ell^{\ast}}(x^{\ast})
  \longmapsto
  \rfield_{\ell^{\ast}}(x^{\ast})+\epsilon J
  \]
  and propagate through the model\;

  Measure
  \[
  \delta\rfield_\ell(x)
  =
  \rfield^{\mathrm{patched}}_\ell(x)
  -
  \rfield^{0}_\ell(x)
  \]
  for all layers $\ell$ and tokens $x$\;

  Remap each response to
  \[
  (\Delta\ell,\Delta x)
  =
  (\ell-\ell^{\ast},x-x^{\ast})
  \]
  and retain only $\Delta\ell\ge0,\Delta x\ge0$\;

  Accumulate $\|\delta\rfield_\ell(x)\|$ in the corresponding
  $(\Delta\ell,\Delta x)$ bin\;
}
\Return the per-bin mean
$\langle\|\delta\rfield\|\rangle(\Delta\ell,\Delta x)$\;
\end{algorithm}

\section{Transformer-field Propagation and Approximate Composition}
\label{sec:e3}

\begin{figure*}[t]
\centering
 \includegraphics[width=0.80\textwidth]{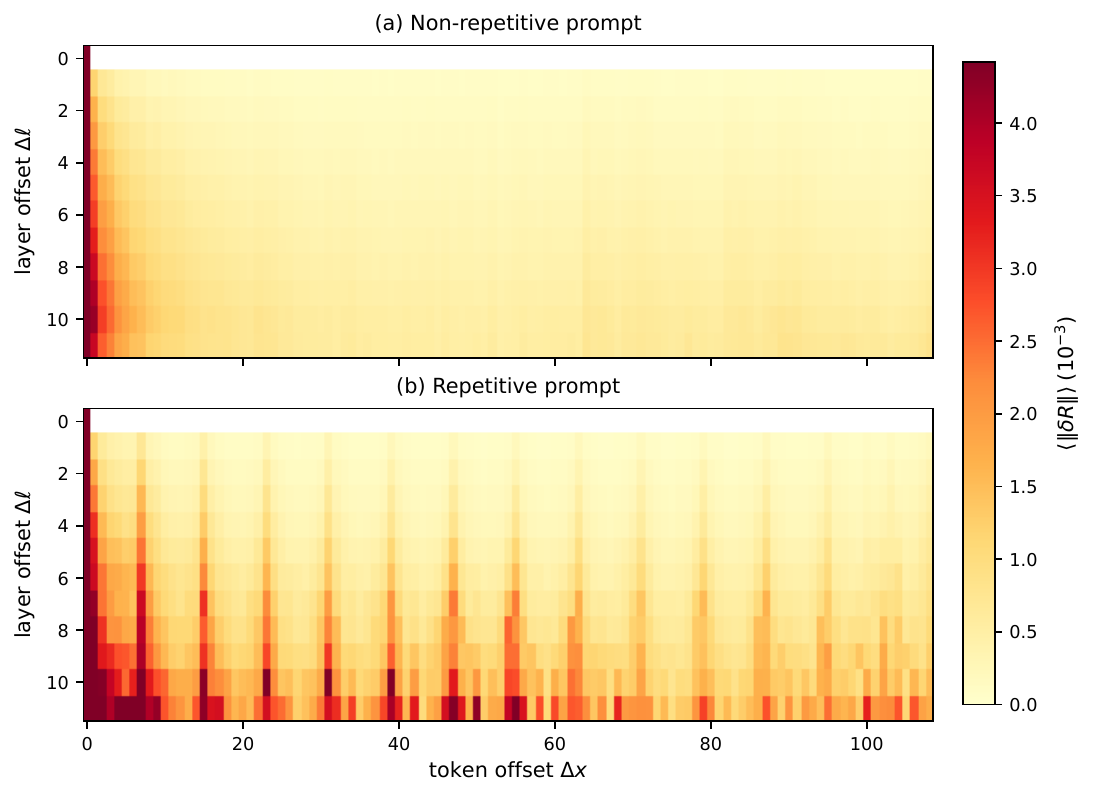}
\caption{\textbf{Empirical Transformer-field response norm in relative coordinates.} Localized Transformer-field perturbations induce downstream Transformer-field displacements $\delta\rfield$ for nonrepetitive and repetitive prompts. The plotted quantity is the mean norm of the induced response $\langle|\delta\rfield|\rangle$, averaged over source sites and expressed in relative coordinates $(\Delta\ell, \Delta x)$. The shared color scale is clipped at the 99.5th percentile of positive off-token values, excluding the $\Delta x=0$ column for scale estimation while still displaying it. The response is concentrated near $\Delta x = 0$ but broadens with increasing $\Delta\ell$, consistent with anisotropic propagation through the Transformer field. Lower-amplitude off-token structure is consistent with nonlocal transport mediated by attention interactions.}
\label{fig:e3_field}
\end{figure*}

Beyond scalar prediction, the response framework describes how Transformer-field perturbations propagate through the model. After a localized additive intervention at the block-output value of the Transformer field $(\ell^{\ast},x^{\ast})$, we measure the Transformer-field response
\begin{equation}
\delta\rfield_\ell(x)
=
\rfield_\ell^{\mathrm{patched}}(x)
-
\rfield_\ell^{0}(x),
\qquad
\ell\ge \ell^{\ast}.
\label{eq:e3_field}
\end{equation}
The norm $\|\delta\rfield_\ell(x)\|$ gives a discrete, direction-specific estimate of the downstream Green-operator propagation from the source site to later layer-token sites.
Algorithm~\ref{alg:field} gives the measurement procedure. In the source-site sweep, one shared unit direction $J$ is used across source positions, and responses are averaged after remapping to relative coordinates $(\Delta\ell,\Delta x)=(\ell-\ell^{\ast},x-x^{\ast})$ over the causal cone $\Delta\ell\ge0,\Delta x\ge0$. Figure~\ref{fig:e3_field} shows the resulting Transformer-field response norm in relative coordinates.

Across the two prompts, the response is strongest near the source site and spreads anisotropically across depth and token offset, with the repetitive prompt producing a periodic response pattern that matches the prompt period.

\begin{figure}[!htbp]
\centering
\includegraphics[width=\linewidth]{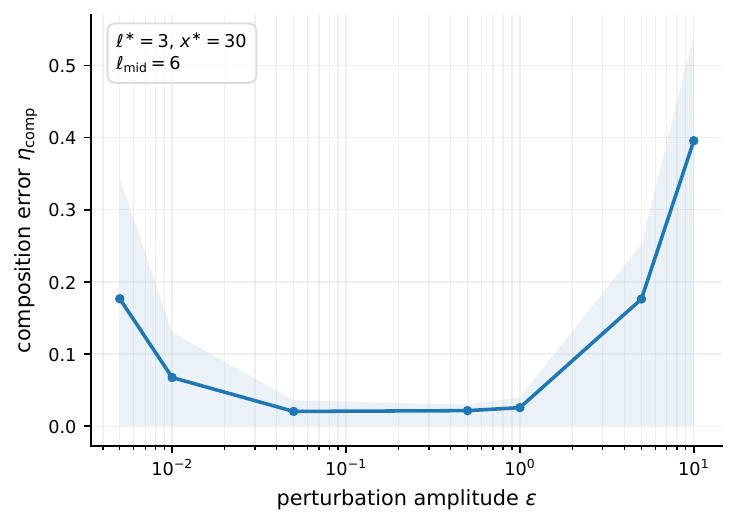}
\caption{\textbf{Approximate depth-composition of Transformer-field responses.}
A directly measured downstream response is compared with a response obtained through an intermediate layer.
The composition error $\eta_{\mathrm{comp}}$ is plotted against perturbation amplitude $\epsilon$.
The curve shows the median over sites and the shaded band shows the interquartile range.
The in-panel annotation reports the source site $(\ell^\ast,x^\ast)$ and the hand-off layer $\ell_{\mathrm{mid}}$.
Errors are lowest at intermediate amplitudes and increase when the perturbation becomes too small to measure robustly or too large for linear transport.}
\label{fig:e3_composition}
\end{figure}

The same Transformer-field response supports a depth-composition test. In the ideal linearized theory, propagation from $\ell_0$ to $\ell_2$ through an intermediate layer $\ell_1$ factors approximately as
\begin{equation}
G(\ell_2;\ell_0)
\approx
G(\ell_2;\ell_1)G(\ell_1;\ell_0),
\qquad
\ell_0<\ell_1<\ell_2 .
\label{eq:e3_composition}
\end{equation}
At the operator level, this motivates the formal composition error for the Green operator
\begin{equation}
\eta_{\mathrm{comp}}^{G}
:=
\frac{
\big\|
\widehat{G}(\ell_2;\ell_0)
-
\widehat{G}(\ell_2;\ell_1)\widehat{G}(\ell_1;\ell_0)
\big\|
}{
\big\|
\widehat{G}(\ell_2;\ell_0)
\big\|
+
\varepsilon_0
}.
\label{eq:e3_eta_green}
\end{equation}
Empirically, we test the same transport principle at the level of measured Transformer-field responses rather than by constructing the three full Green operators. For a source perturbation at $(\ell^{\ast},x^{\ast})$, we compare the directly measured downstream field with the field obtained by re-injecting the intermediate response at a hand-off layer $\ell_{\mathrm{mid}}$ and propagating forward. The field-level composition error is
\begin{equation}
\eta_{\mathrm{comp}}^{\rfield}
:=
\frac{
\big\|
\delta\rfield^{\mathrm{direct}}_{\ell>\ell_{\mathrm{mid}}}
-
\delta\rfield^{\mathrm{reprop}}_{\ell>\ell_{\mathrm{mid}}}
\big\|
}{
\big\|
\delta\rfield^{\mathrm{direct}}_{\ell>\ell_{\mathrm{mid}}}
\big\|
+
\varepsilon_0
}.
\label{eq:e3_eta_comp}
\end{equation}
Thus Eq.~\eqref{eq:e3_eta_green} is the ideal Green-operator criterion, while Eq.~\eqref{eq:e3_eta_comp} is the empirical Transformer-field realization. Algorithm~\ref{alg:green} gives this direct-versus-repropagated composition test. Figure~\ref{fig:e3_composition} summarizes its amplitude dependence; the plotted $\eta_{\mathrm{comp}}$ denotes the field-level composition error $\eta_{\mathrm{comp}}^{\rfield}$ of Eq.~\eqref{eq:e3_eta_comp}.

The small composition error in the intermediate-$\epsilon$ range shows that the measured hand-off field at $\ell_{\mathrm{mid}}$ retains enough information to reproduce the downstream response, while the increase at large $\epsilon$ marks the breakdown of this approximate linear transport.

Together, Figs.~\ref{fig:e3_field} and~\ref{fig:e3_composition} show Transformer-field propagation in the finite model: localized Transformer-field perturbations produce structured downstream Transformer-field responses, and those responses  can be approximately handed off through an intermediate layer within the perturbative regime.

\section{Sensitivity Concentration and Direction Families}
\label{sec:e5}

\begin{figure*}[t]
\centering
\includegraphics[width=0.92\textwidth]{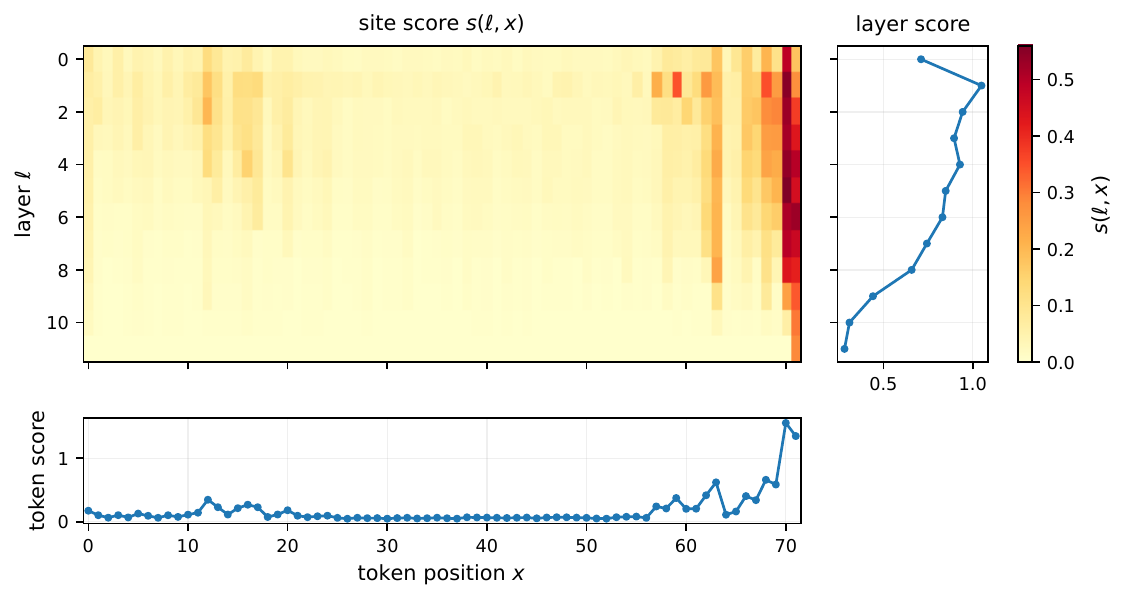}
\caption{\textbf{Sensitivity site score over layer-token sites.} The heat map shows the site score $s(\ell,x)=\|a(\ell,x)\|=\|\partial y/\partial \rfield_\ell(x)\|$, where $a(\ell,x)$ is the sensitivity field of the scalar observable. The bottom token score is the root-sum-square marginal over layers, and the right layer score is the root-sum-square marginal over tokens. High-score sites are localized in the depth-token field, indicating where the scalar observable is most sensitive to local Transformer-field perturbations.}
\label{fig:e5_concentration}
\end{figure*}

\begin{algorithm}[!t]
\caption{Empirical composition test through an intermediate layer}
\label{alg:green}
\KwIn{Transformer-field evolution; query $q$; source site $(\ell^{\ast},x^{\ast})$;
      hand-off layer $\ell_{\mathrm{mid}}$; direction $J$, $\|J\|=1$;
      amplitude $\epsilon$; floor $\varepsilon_0$}
\KwOut{field-level composition error $\eta_{\mathrm{comp}}^{\rfield}$}

Compute the clean Transformer field $\rfield^{0}_{\ell}(x)$ for the query $q$\;

\BlankLine
\textbf{Direct response.}\;
Apply
\[
\rfield_{\ell^{\ast}}(x^{\ast})
\longmapsto
\rfield_{\ell^{\ast}}(x^{\ast})+\epsilon J
\]
and measure
\[
\delta\rfield^{\mathrm{direct}}_{\ell}(x)
=
\rfield^{\mathrm{direct}}_{\ell}(x)
-
\rfield^{0}_{\ell}(x).
\]

\BlankLine
\textbf{Hand-off field.}\;
Extract
\[
f^{\mathrm{mid}}(x)
=
\delta\rfield^{\mathrm{direct}}_{\ell_{\mathrm{mid}}}(x).
\]
For the linear reconstruction, $f^{\mathrm{mid}}$ is the first-order response at
$\ell_{\mathrm{mid}}$ induced by $\epsilon J$\;

\BlankLine
\textbf{Re-propagated response.}\;
Starting again from the clean query, impose
\[
\rfield_{\ell_{\mathrm{mid}}}(x)
\longmapsto
\rfield_{\ell_{\mathrm{mid}}}(x)+f^{\mathrm{mid}}(x)
\]
and measure
\[
\delta\rfield^{\mathrm{reprop}}_{\ell}(x)
=
\rfield^{\mathrm{reprop}}_{\ell}(x)
-
\rfield^{0}_{\ell}(x).
\]

\BlankLine
\textbf{Composition error.}\;
\[
\eta_{\mathrm{comp}}^{\rfield}
=
\frac{
\big\|
\delta\rfield^{\mathrm{direct}}_{\ell>\ell_{\mathrm{mid}}}
-
\delta\rfield^{\mathrm{reprop}}_{\ell>\ell_{\mathrm{mid}}}
\big\|
}{
\big\|
\delta\rfield^{\mathrm{direct}}_{\ell>\ell_{\mathrm{mid}}}
\big\|
+
\varepsilon_0
}.
\]

\Return $\eta_{\mathrm{comp}}^{\rfield}$\;
\end{algorithm}

The sensitivity field $a(\ell,x)$ compresses scalar responses to patching, but it still ranges over all layer-token sites. We therefore ask whether the response is concentrated on a small set of sites. We score each site by
\begin{equation}
s(\ell,x)
:=
\|a(\ell,x)\|.
\label{eq:e5_score}
\end{equation}
For a fixed site, the maximally aligned first-order direction is proportional to $a(\ell,x)$ itself. This is the local version of the patch-inference rule: in the linear regime, the direction that maximizes scalar response aligns with the sensitivity. Random local directions, random global directions, and gradient-aligned directions serve as controls for separating sensitivity alignment from generic Transformer-field norm injection.

Figure~\ref{fig:e5_concentration} visualizes the site score $s(\ell,x)=\|a(\ell,x)\|$ over the layer-token lattice. The heat map and marginals show that the scalar-output sensitivity is highly nonuniform, with most of the mass concentrated on a limited set of layer-token sites. Thus $a(\ell,x)$ does more than predict local scalar responses: its norm gives a site-selection rule for reducing the intervention space, while the local gradient direction gives the aligned perturbation direction at the selected sites.

\begin{figure}[t]
\centering
\includegraphics[width=\linewidth]{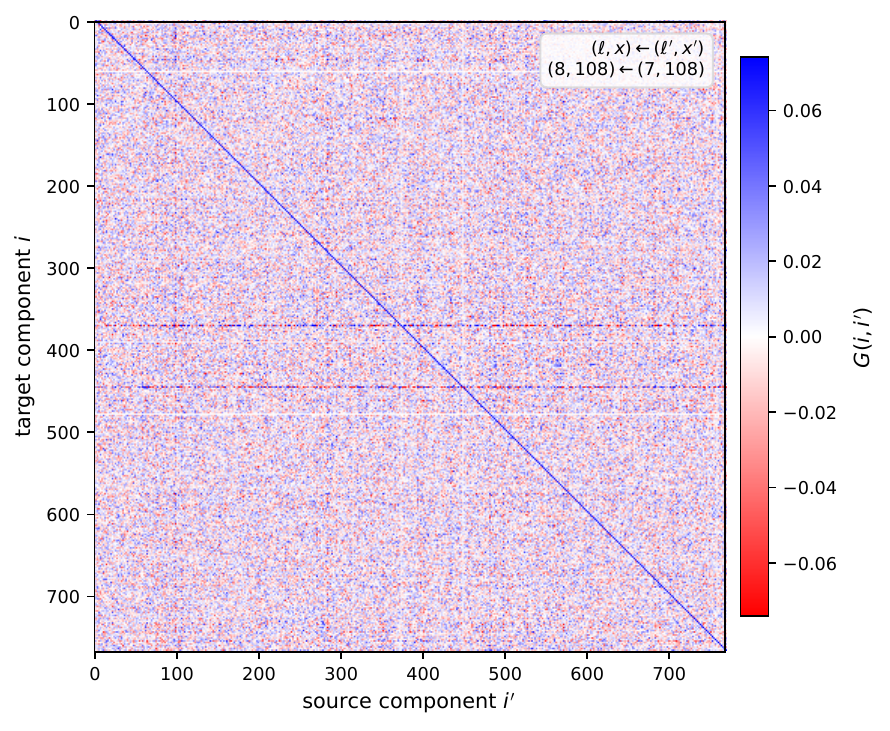}
\caption{\textbf{Component-level view of a sliced Green operator.} For a fixed same-token, adjacent-layer source-target pair, the site-to-site sliced Green operator $G_{(\ell,x)\leftarrow(\ell',x')}$ is plotted over source and target Transformer-field components. The in-panel annotation reports the fixed target and source sites $(\ell,x)\leftarrow(\ell',x')$. The dominant diagonal structure shows componentwise persistence through the Transformer field, while off-diagonal structure records component mixing induced by the Transformer block between the two sites.}
\label{fig:e6_green_slice_matrix}
\end{figure}

\begin{figure}[t]
\centering
\includegraphics[width=\linewidth]{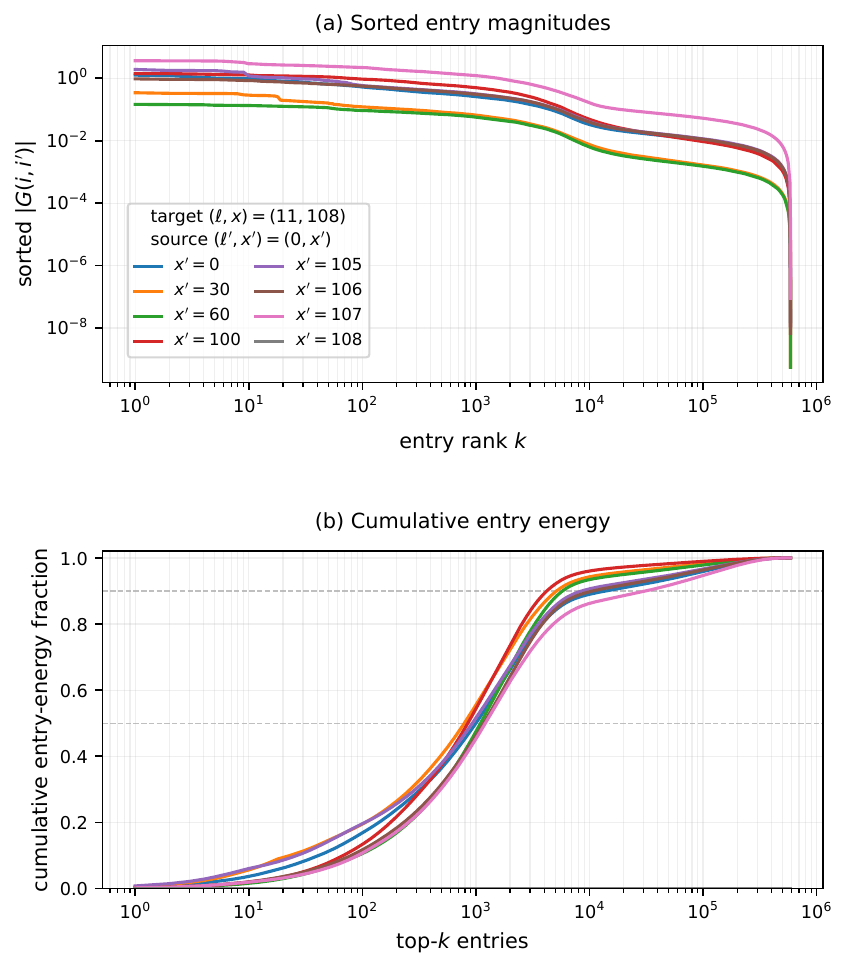}
\caption{\textbf{Entrywise concentration of sliced Green operators.} For a fixed target site $(\ell,x)=(11,108)$ and source layer $\ell'=0$, the component entries of site-to-site sliced Green operators $G_{(\ell,x)\leftarrow(\ell',x')}$ are analyzed across source-token positions $x'$. Panel (a) shows sorted entry magnitudes $|G(i,i')|$. Panel (b) shows the cumulative entry-energy fraction, where entry energy is proportional to $|G(i,i')|^2$. The cumulative curves show that response energy is spread across many entries, limiting simple magnitude-based compression.}
\label{fig:e6_slices}
\end{figure}

\section{Reduced Response Descriptions and Sliced Green Operators}
\label{sec:e6}

A full Green operator over layers, tokens, and Transformer-field components has formal size
\begin{equation}
(L\cdot N_x\cdot d_{\mathrm{model}})^2,
\label{eq:e6_size}
\end{equation}
for $L$ layers, $N_x$ tokens, and Transformer-field dimension $d_{\mathrm{model}}$. Even at GPT-2 scale, this object is too large to estimate exhaustively.

We therefore study reduced representations of the Green operator: observable-specific sensitivity site scores, which select source sites, and component-level sliced Green operators, which describe Transformer-field propagation between fixed layer-token sites.

\subsection{Sliced Green operators}

Instead of estimating the full Green operator $G(\ell,x,i;\ell',x',i')$, we fix source and target layer-token sites and estimate the component-level sliced Green operator

\begin{equation}
G_{(\ell,x)\leftarrow(\ell',x')}(i,i')
:=
\frac{\partial \rfield_\ell(x,i)}
{\partial \rfield_{\ell'}(x',i')}
\in
\mathbb{R}^{d_{\mathrm{model}}\times d_{\mathrm{model}}}.
\label{eq:e6_slice}
\end{equation}

This sliced Green operator is the differential of the target Transformer field with respect to a localized source Transformer-field perturbation, evaluated at the clean trajectory. Its component structure can be summarized by the matrix pattern itself, entrywise concentration, cumulative Frobenius energy, and variation across source-token positions.

Figures~\ref{fig:e6_green_slice_matrix} and~\ref{fig:e6_slices} are the two empirical views of these sliced Green operators. Figure~\ref{fig:e6_green_slice_matrix} displays one component-level sliced Green operator as a matrix over source and target Transformer-field components, making diagonal persistence and off-diagonal mixing visible directly. Figure~\ref{fig:e6_slices} then asks how concentrated the entries of such sliced Green operators are by sorting component magnitudes and plotting cumulative Frobenius energy. The pair therefore connects the local sliced-Green-operator definition in Eq.~\eqref{eq:e6_slice} to the practical claim that useful site-to-site response structure within the Transformer field can be
studied without constructing the full Green operator.

The empirical result is that Green-operator structure survives strong reduction. Sensitivity site scores identify a reduced set of observable-specific candidate intervention sites, while sliced Green operators reveal local site-to-site geometry without constructing the full Green operator. These reduced objects expose response strength, component mixing, entrywise concentration, and variation across source-token positions in a computationally tractable form.

\section{Prompt-Induced Transformer-Field Displacements as Transformations}
\label{sec:e7}

\begin{figure}[!t]
\centering
\includegraphics[width=\linewidth]{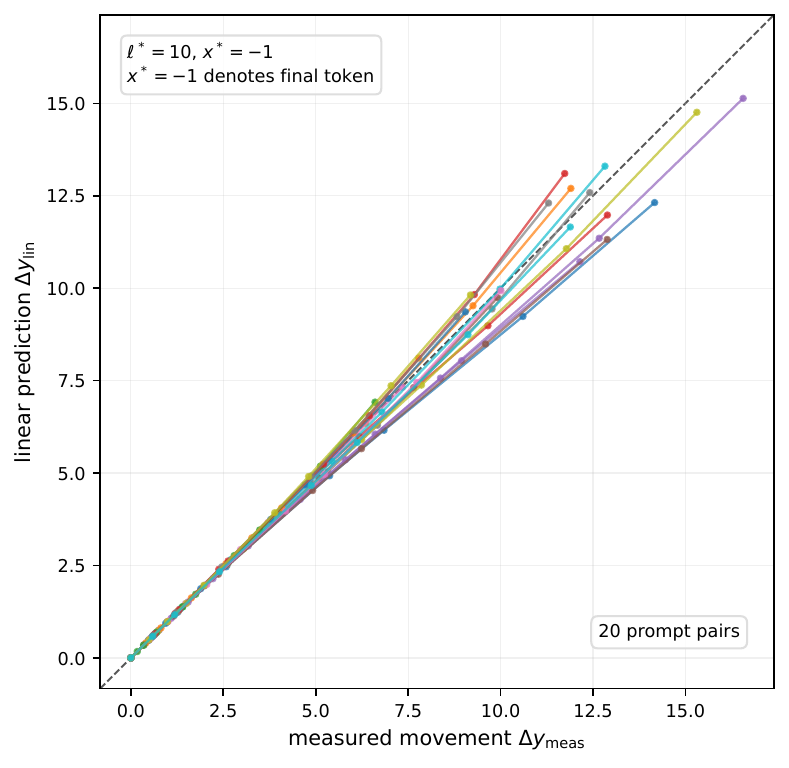}
\caption{\textbf{Sensitivity prediction for prompt-induced Transformer-field displacements.} For 20 country--capital prompt pairs, the Transformer-field displacement $J_{q_A\to q_B}$ is computed at $(\ell^\ast=10,x^\ast=-1)$ in GPT-2 small, with $x^\ast=-1$ denoting the final token, and injected into the Transformer field of the $q_A$ run with amplitude $\epsilon$. The plot compares the first-order prediction $\Delta y_{\rm lin}$ with the measured movement $\Delta y_{\rm meas}$ across pairs and amplitudes. Each colored line corresponds to one country--capital pair, with points along the line showing the swept amplitudes $\epsilon$. Alignment with the diagonal indicates that the displacement effect is predicted by the same local sensitivity rule used for Transformer-field patching.}
\label{fig:e7_prompt_transform}
\end{figure}

\begin{figure}[!t]
\centering
\includegraphics[width=\linewidth]{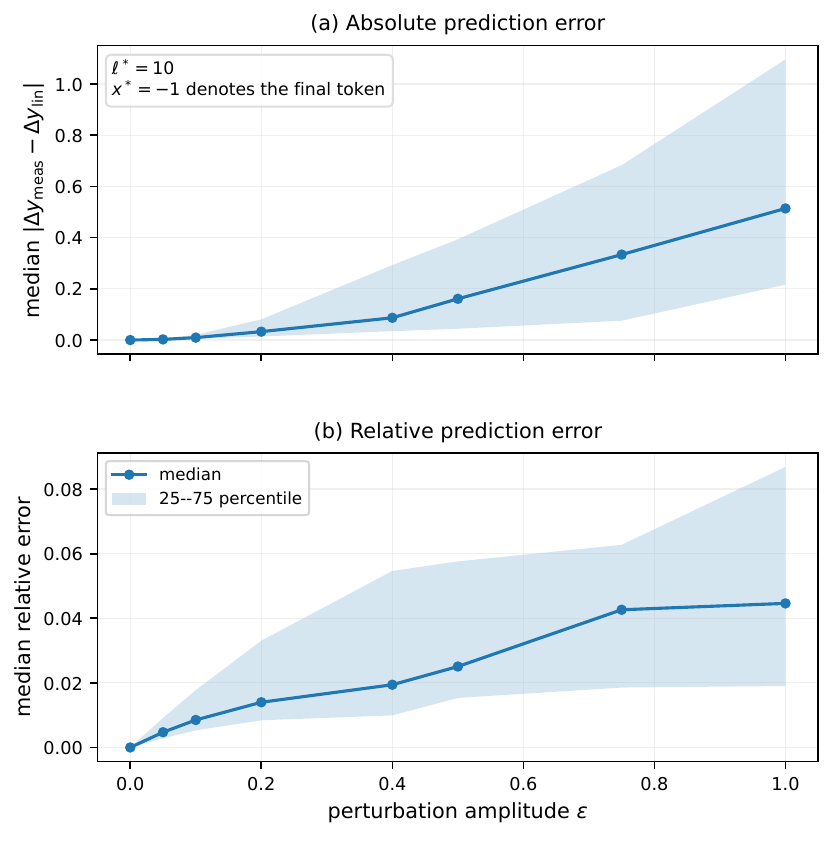}
\caption{\textbf{Linear-response error for prompt-induced Transformer-field displacements.} At the same intervention site as Fig.~\ref{fig:e7_prompt_transform}, absolute and relative errors between $\Delta y_{\rm meas}$ and $\Delta y_{\rm lin}$ are computed for each prompt pair and summarized as a function of $\epsilon$. The solid line shows the median across the 20 prompt pairs, and the shaded band indicates the 25th--75th percentile range. Absolute error increases with amplitude, while the median relative error remains small over the tested range.}
\label{fig:e7_linear_error}
\end{figure}

The same response framework applies to transformations between prompts. Given two prompts $q_A$ and $q_B$, define the local Transformer-field displacement at a fixed layer-token site by
\begin{equation}
J_{q_A\to q_B}
:=
\rfield(q_B;\ell^{\ast},x^{\ast})
-
\rfield(q_A;\ell^{\ast},x^{\ast})
\label{eq:e7_prompt_direction}
\end{equation}
Injecting $J_{q_A\to q_B}$ into the Transformer field of the $q_A$ run tests whether the Transformer-field difference between prompts acts as a local transformation direction. This is the difference-of-activations construction behind task vectors,\cite{hendel2023taskvectors} function vectors,\cite{todd2024functionvectors} and activation addition / representation engineering,\cite{turner2023activation,zou2023representation} here interpreted as a localized source in the response formalism.

For correct answer tokens $A$ and $B$ associated with prompts $q_A$ and $q_B$, define the pairwise logit-difference observable
\begin{equation}
y_{A\to B}(q)
:=
\mathrm{logit}(B\mid q)-\mathrm{logit}(A\mid q)
.
\label{eq:e7_pairwise_logit}
\end{equation}
The clean $A$-to-$B$ shift is
\begin{equation}
\Delta_{A\to B}
:=
y_{A\to B}(q_B)-y_{A\to B}(q_A)
,
\label{eq:e7_clean_shift}
\end{equation}
The corresponding local sensitivity at the source run is
\begin{equation}
a_{q_A}(\ell^\ast,x^\ast)
:=
\frac{\partial y_{A\to B}}
{\partial \rfield(q_A;\ell^\ast,x^\ast)}
.
\label{eq:e7_source_sensitivity}
\end{equation}
We measure the normalized patch effect by the toward fraction
\begin{equation}
f_{A\to B}(\epsilon)
:={}
\frac{
\delta y_{A\to B}[\epsilon J_{q_A\to q_B}]
}{
\mathrm{sign}(\Delta_{A\to B})
\max(|\Delta_{A\to B}|,\varepsilon_0)
}
\label{eq:e7_toward_fraction}
\end{equation}
Here $\delta y_{A\to B}$ denotes the patch-induced change in $y_{A\to B}$.
The denominator uses $\Delta_{A\to B}$ as the normalization scale while preserving its sign. Thus $f_{A\to B}(\epsilon)$ measures the fraction of the clean $A$-to-$B$ logit-difference shift produced by injecting $\epsilon J_{q_A\to q_B}$ into the Transformer field of the $q_A$ run.

\begin{figure}[!t]
\centering
\includegraphics[width=\linewidth]{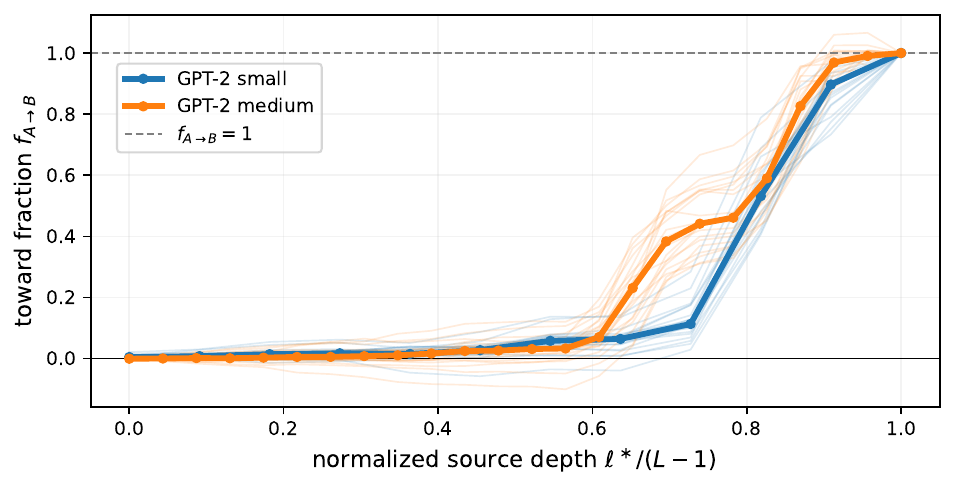}
\caption{\textbf{Layer dependence of prompt-induced Transformer-field displacement effects.} For each prompt pair, $J_{q_A\to q_B}$ is injected at the final token with $\epsilon=1$ while the source layer $\ell^\ast$ is swept. The horizontal axis shows normalized source depth $\ell^\ast/(L-1)$, allowing GPT-2 small and GPT-2 medium to be compared on a common depth scale. Thin lines show individual prompt pairs and thick lines show the median. The dashed line marks $f_{A\to B}=1$, corresponding to the full clean prompt-induced logit-difference shift. Both models show weak transfer at early layers and a sharp rise near the end of the network.}
\label{fig:e7_layer_sweep}
\end{figure}

\begin{figure}[!t]
\includegraphics[width=\linewidth]{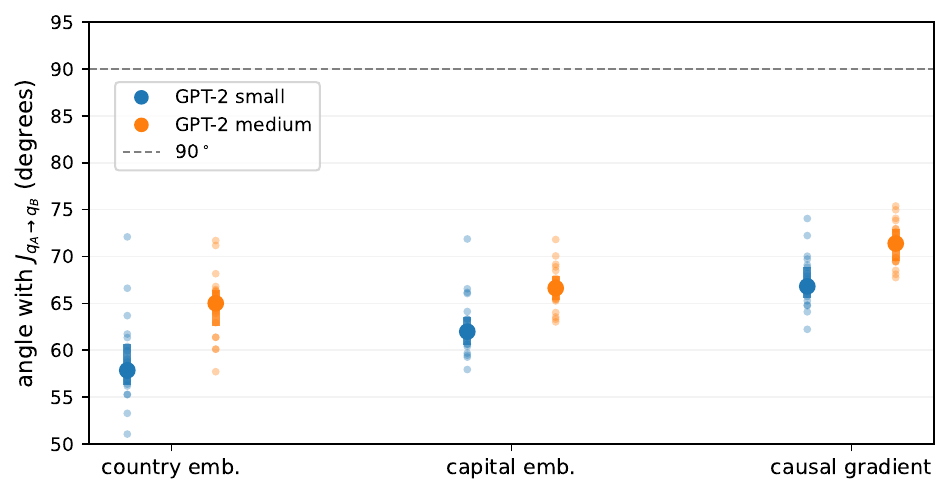}
\caption{\textbf{Alignment of prompt-induced Transformer-field displacement directions.}
For each prompt pair, the Transformer-field displacement $J_{q_A\to q_B}$ is compared with three reference directions: the country embedding difference, the capital embedding difference, and the local causal gradient $a_{q_A}$. Points show individual prompt pairs, and larger markers show the median for each model and reference direction. Angles are measured at the same final-token sites used for the local tests: $(\ell^\ast,x^\ast)=(10,-1)$ in GPT-2 small and $(22,-1)$ in GPT-2 medium, where $x^\ast=-1$ denotes the final token. Angles below $90^\circ$ indicate positive alignment with the corresponding reference direction. In both GPT-2 small and GPT-2 medium, the displacement direction exhibits consistent positive alignment with semantic embedding differences and with the local causal gradient.}
\label{fig:e7_direction_angles}
\end{figure}

\begin{figure*}[!t]
\centering
\includegraphics[width=0.80\textwidth]{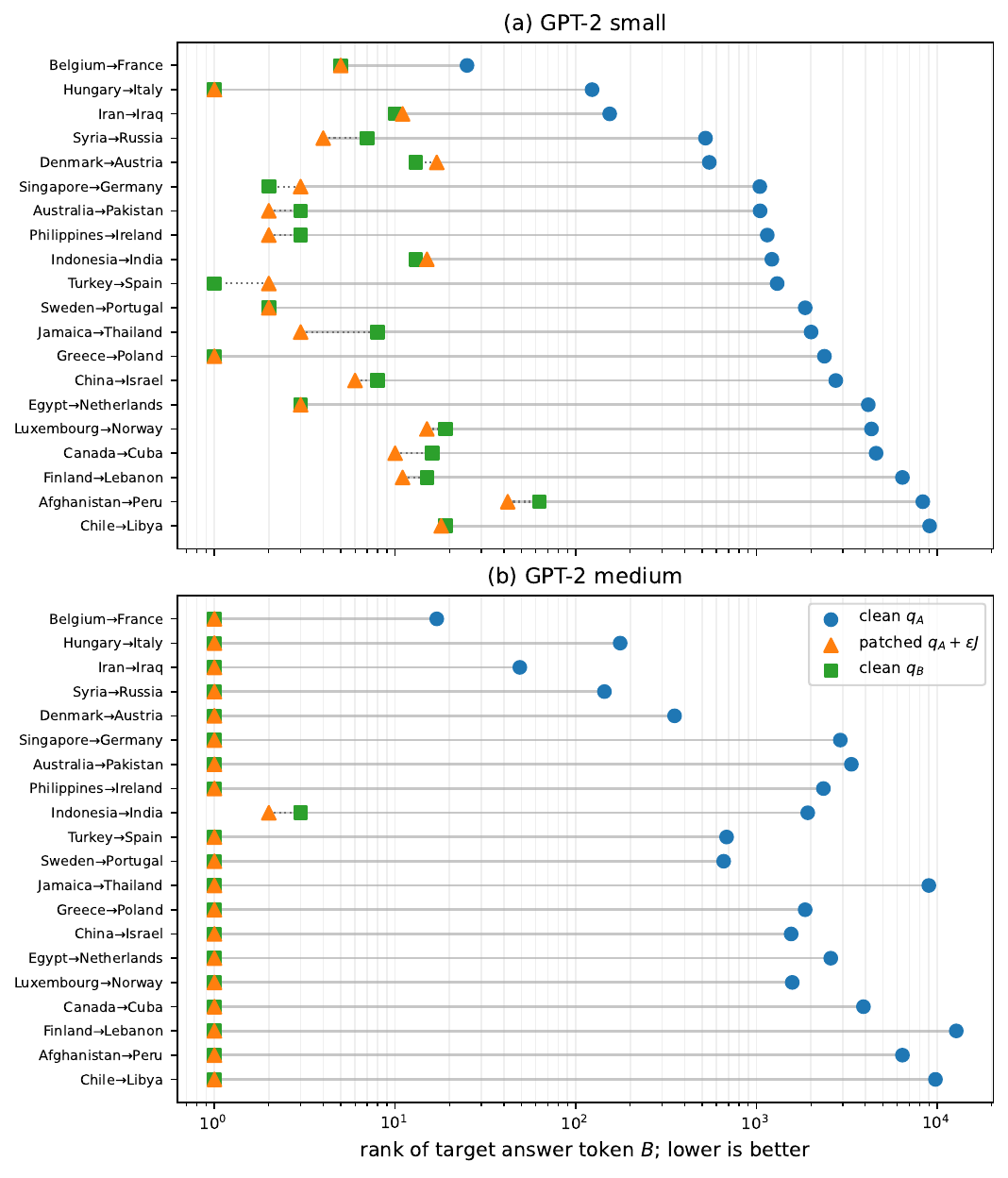}
\caption{\textbf{Target-answer rank after Transformer-field displacement patching.}
For each prompt pair, the vocabulary rank of the target answer token $B$ is shown for the clean source prompt $q_A$, the source run patched with $\epsilon J_{q_A\to q_B}$ at $\epsilon=1$, and the clean target prompt $q_B$. Lower rank is better; on the log-scaled axis, $10^0=1$ corresponds to a top-ranked answer. (a) GPT-2 small. The patch consistently moves the target answer toward rank one and often matches or improves upon the clean target-prompt rank. (b) GPT-2 medium. The effect is stronger: for most prompt pairs, the patched rank is nearly identical to the clean target-prompt rank and frequently reaches rank one.}
\label{fig:e7_answer_ranks}
\end{figure*}

The experiment uses matched country--capital prompt pairs with the same template, such as $q_A={}$``The capital of Spain is'' and $q_B={}$``The capital of France is'', with correct answer tokens $A={}$``Madrid'' and $B={}$``Paris''. For each pair, the displacement $J_{q_A\to q_B}$ is measured at the final token of the source prompt $q_A$. Because the prompts differ only in the country entity, this displacement is treated as a controlled country-to-country transformation direction. Figures~\ref{fig:e7_prompt_transform} and~\ref{fig:e7_linear_error} first test whether the scalar effect of this displacement is predicted by the local sensitivity. Figure~\ref{fig:e7_layer_sweep} then sweeps the source layer to locate where the displacement is effective. Figure~\ref{fig:e7_direction_angles} compares the displacement direction with the causal gradient and embedding-space reference directions. Finally, Fig.~\ref{fig:e7_answer_ranks} asks whether the same local Transformer-field displacement also changes the full-vocabulary rank of the target answer.

Together, Figs.~\ref{fig:e7_prompt_transform} and~\ref{fig:e7_linear_error} validate the use of prompt-induced Transformer-field displacements as local sources within the same first-order response regime used throughout the paper. Figure~\ref{fig:e7_prompt_transform} shows that the measured scalar response is well predicted by the linear-response estimate $\Delta y_{\rm lin}=\epsilon\, a_{q_A}\!\cdot\!J_{q_A\rightarrow q_B}$ across diverse country--capital prompt pairs and intervention amplitudes, with most trajectories remaining close to the diagonal. Figure~\ref{fig:e7_linear_error} further shows that the associated prediction errors remain controlled across the tested amplitude range. Additional control directions generally produce weaker and less consistent targetward movements than the measured Transformer-field displacement direction $J_{q_A\rightarrow q_B}$, with many controls remaining near zero or exhibiting mixed-sign responses. Taken together, these results indicate that prompt-induced Transformer-field displacements obey the same local sensitivity rule previously validated for Transformer-field patching.

The layer sweep in Fig.~\ref{fig:e7_layer_sweep} localizes the transformation effect in depth. The prompt displacement is not equally effective at all Transformer-field sites; it becomes useful mainly late in the computation. When plotted against normalized depth, GPT-2 small and GPT-2 medium exhibit qualitatively similar transfer profiles, with weak transfer at early depths and a sharp rise near the end of the network.

The direction analysis in Fig.~\ref{fig:e7_direction_angles} places the Transformer-field displacement in relation to both causal and embedding-space directions. The country and capital embedding differences provide semantic reference directions for the prompt change, while the local causal gradient $a_{q_A}$ gives the direction that most directly increases the measured logit-difference response. The displacement has positive alignment with the country and capital embedding-space references and with the causal-gradient reference, indicating that it is aligned with both semantic and causal directions in the Transformer field.

The answer-rank plot in Fig.~\ref{fig:e7_answer_ranks} tests whether the same local Transformer-field displacement changes the full-vocabulary ordering. The previous figures measure scalar response and geometry; this rank plot shows that the target answer can move upward in the vocabulary ranking after patching. In GPT-2 small, the patched source run often ranks the target token as well as or better than the clean target prompt, although the effect is not uniform across all pairs. In GPT-2 medium, the effect is stronger: for most prompt pairs, the patched rank is nearly identical to the clean target-prompt rank and frequently reaches rank one.

Taken together, Figs.~\ref{fig:e7_prompt_transform}--\ref{fig:e7_answer_ranks} show that prompt-induced Transformer-field displacements can act as local transformation directions. Their scalar effects are predicted by the causal sensitivity, their effectiveness is concentrated near the end of the network, their directions align with semantic and causal references, and their patches improve target-answer ranks across the tested prompt pairs.

\section{Discussion and Conclusions}
\label{sec:discussion}

We have developed a continuous-depth field-theoretic framework for Transformer-field patching and tested its linear-response objects in GPT-2-style models. The Transformer field is defined over depth and token position; patching becomes localized source insertion; patch effects are predicted by sensitivity fields; downstream Transformer-field propagation is measured through Transformer-field responses and sliced Green operators; and an adjoint action principle frames patch-site inference as an inverse problem over sources \(J\).

The empirical results show that these response objects provide a practical language for organizing Transformer-field interventions. Local perturbations exhibit a bounded local linear regime; patch effects are predicted by sensitivity fields in the local linear regime; localized Transformer-field interventions generate structured downstream Transformer-field responses; high-sensitivity sites concentrate observable influence; sliced Green operators expose local site-to-site geometry; and prompt-induced Transformer-field displacements behave as pair-specific transformation directions in controlled prompt-pair experiments.

The conceptual point is that these measurements are finite-dimensional instances of one response formalism. The sensitivity field \(a(\ell,x)\) is the adjoint response to the output observable, while Green functions and their discrete Green-operator realizations describe forward transport through the Transformer-field dynamics. Together they recast patching as an operator-based description of interventions over the depth-token Transformer field, rather than as a collection of isolated causal probes.

The broader purpose of the formalism is patch-site inference. In the local linear regime, the Green-function equation supplies the forward operator of a constrained inverse problem: given a desired behavioral or Transformer-field shift, infer a source \(J\) whose propagated response realizes the target. This paper establishes the field-theoretic formulation and shows empirically that the corresponding linear-response objects are measurable, predictive, and structured in finite Transformer models. It also motivates a scaling hypothesis: if related models approximate a shared latent response geometry, then response information measured in a smaller model may help define candidate patch-site maps in a larger model, supplemented by local anchors and sparse validation. Thus the contribution is both theoretical and empirical: it provides a principled route from mechanistic localization to inferred interventions, and from single\mbox{-}model response measurements toward possible model-scale transfer.

\onecolumngrid
\FloatBarrier


\end{document}